\documentclass[lettersize,journal]{IEEEtran}
\usepackage{amsmath,amsfonts}
\usepackage{algorithmic}
\usepackage{algorithm}
\usepackage{array}
\usepackage[caption=false,font=normalsize,labelfont=sf,textfont=sf]{subfig}
\usepackage{textcomp}
\usepackage{stfloats}
\usepackage{url}
\usepackage{verbatim}
\usepackage{graphicx}
\usepackage{cite}
\usepackage{multirow}
\usepackage{xcolor}         
\begin{document}

\title{Interpretable Zero-shot Referring Expression Comprehension with Query-driven Scene Graphs}

\author{Yike~Wu,
 Necva~Bolucu,
 Stephen~Wan,
 Dadong~Wang,
 Jiahao~Xia,
 Jian Zhang
\IEEEcompsocitemizethanks{\IEEEcompsocthanksitem Yike Wu, Jiahao Xia, Jian Zhang are with the Faculty of Engineering and IT, University of Technology Sydney,
NSW 2007, Australia. E-mail: Yike.Wu@student.uts.edu.au, \{Jiahao.Xia-1, Jian.Zhang\}@uts.edu.au\
\IEEEcompsocthanksitem Necva~Bolucu, Stephen~Wan, and Dadong~Wang are with Data61, Commonwealth Scientific and Industrial Research Organisation, NSW 2122, Australia. E-mail: \{Necva.Bolucu, Stephen.Wan, Dadong.Wang\}@data61.csiro.au
\IEEEcompsocthanksitem Corresponding author: Jian Zhang \protect\\
}}

\markboth{Journal of \LaTeX\ Class Files,~Vol.~14, No.~8, August~2021}%
{Shell \MakeLowercase{\textit{et al.}}: A Sample Article Using IEEEtran.cls for IEEE Journals}


\maketitle

\begin{abstract}
Zero-shot referring expression comprehension (REC) aims to locate target objects in images given natural language queries without relying on task-specific training data, demanding strong visual understanding capabilities. Existing Vision-Language Models~(VLMs), such as CLIP, commonly address zero-shot REC by directly measuring feature similarities between textual queries and image regions. However, these methods struggle to capture fine-grained visual details and understand complex object relationships. Meanwhile, Large Language Models~(LLMs) excel at high-level semantic reasoning, their inability to directly abstract visual features into textual semantics limits their application in REC tasks. To overcome these limitations, we propose \textbf{SGREC}, an interpretable zero-shot REC method leveraging query-driven scene graphs as structured intermediaries. Specifically, we first employ a VLM to construct a query-driven scene graph that explicitly encodes spatial relationships, descriptive captions, and object interactions relevant to the given query. By leveraging this scene graph, we bridge the gap between low-level image regions and higher-level semantic understanding required by LLMs. Finally, an LLM infers the target object from the structured textual representation provided by the scene graph, responding with detailed explanations for its decisions that ensure interpretability in the inference process. Extensive experiments show that SGREC achieves top-1 accuracy on most zero-shot REC benchmarks, including RefCOCO val (\textbf{66.78\%}), RefCOCO+ testB (\textbf{53.43\%}), and RefCOCOg val (\textbf{73.28\%}), highlighting its strong visual scene understanding.
\end{abstract}


\section{Introduction}
Referring expression comprehension~(REC)~\cite{qiao2020referring} involves identifying the target object in an image that best corresponds to a given language query. It serves as the foundation for various real-world applications, including vision-language navigation~\cite{gu2022vision, qiao2023march, zhang2023uni3d} and image captioning~\cite{stefanini2022show}. However, the high cost of annotating query-region pairs limits the scale and diversity of training data, making it challenging for supervised methods to generalize to novel queries and unseen visual objects. Consequently, zero-shot REC, which does not rely on task-specific training data, has gained increased interest in recent years. It offers a practical solution for real-world scenarios where labeled data is limited or unavailable.

\begin{figure}
\centering
\includegraphics[width=0.9\linewidth]{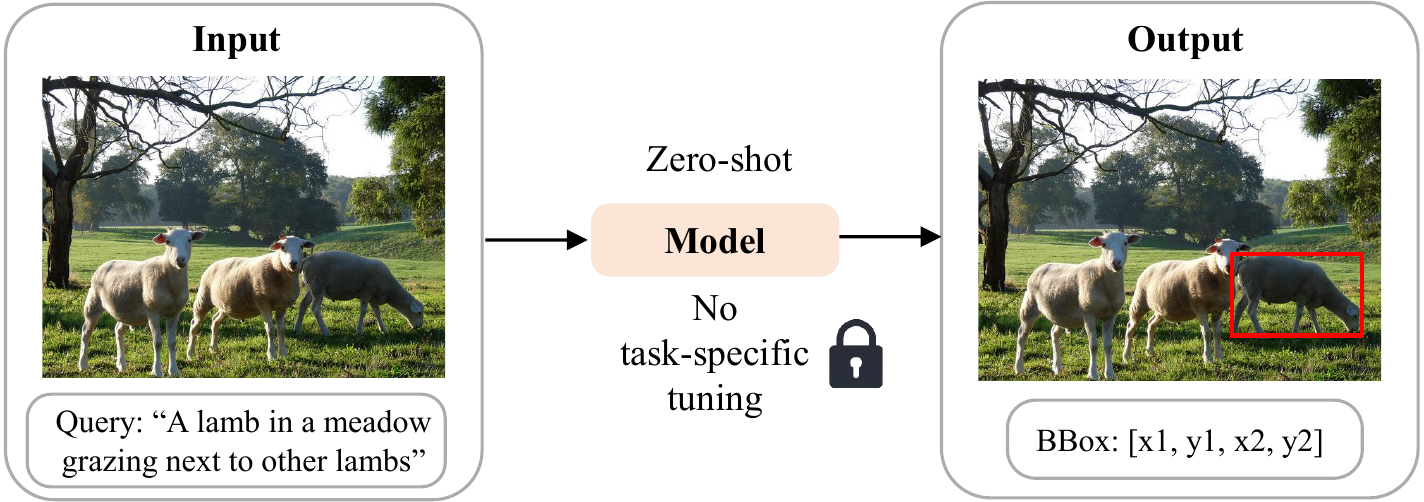}
\caption{Problem definition of zero-shot referring expression comprehension.}
\label{definition}
\end{figure}

Existing works have applied off-the-shelf Vision-Language Models~(VLMs) to zero-shot REC tasks, benefiting from their large-scale vision-language pretraining that enables strong semantic alignment between visual regions and textual descriptions. CLIP~\cite{radford2021learning}, a representative model, identifies the most likely matching object by comparing feature similarities between candidate objects and the given text. Although CLIP-based zero-shot REC methods~\cite{subramanian2022reclip,shtedritski2023does,yang2024fine,qiu2024mcce} have made notable progress, they generally struggle to comprehend contextual information in visual scenes, since CLIP lacks modules dedicated to contextual relationship modeling~\cite{yuksekgonul2022and} and logical reasoning. Furthermore, the diversity of language queries, involving spatial, attribute-based, and semantic relationships, demands a comprehensive understanding of the visual scene, posing additional challenges for zero-shot models.

\begin{figure}[t]
    \centering
    \includegraphics[scale=0.26]{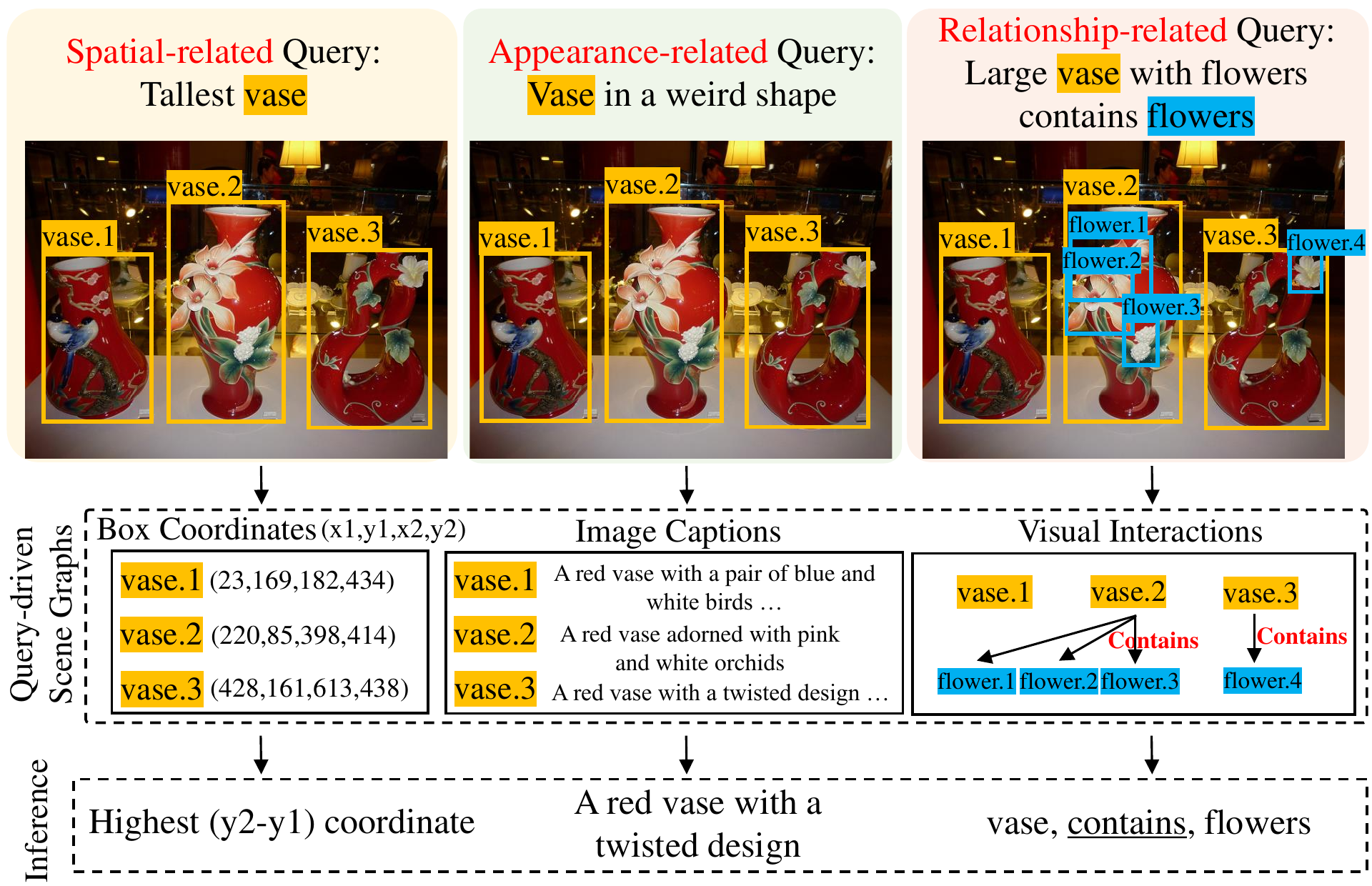}
 \caption{
 Considering the spatial information, object appearance, and semantic relationships within the query, the generated query-driven scene graphs employ the relevant objects' coordinates, captions, and visual interactions to describe visual scenes, facilitating inference by LLMs comprehensively. The queries from the datasets are natural and written in an informal style.
 }
 \label{fig_title}
\end{figure}

To identify target objects, humans tend to interpret queries and analyze relationships between objects, abstracting related image regions into higher-level semantics. This process requires strong visual scene comprehension and inference abilities. As illustrated in Figure~\ref{fig_title}, examples from the RefCOCO/+/g datasets~\cite{yu2016modeling,mao2016generation} highlight queries that require different ways to interpret for accurate localization. These queries can be roughly categorized into spatial-related, appearance-related, and non-spatial relationship-related types. Spatial-related queries focus on modeling relative positional relationships between objects based on their coordinate values, while appearance-related queries depend on analyzing object descriptions to infer the target. Relationship-related queries are longer and more complex and involve incorporating contextual information to represent object interactions in complex scenes. To this end, an explicit representation is needed to clarify how objects relate to each other spatially and semantically, enabling a more accurate understanding of visual scenes.

Scene graphs~\cite{chang2021comprehensive}, structured representations of relationships between objects, provide explicit guidance for representing visual scenes, offering a promising yet underexplored approach for zero-shot REC. In zero-shot REC, which involves complex scenarios with unseen categories and novel relationships, making it challenging to generate meaningful scene graphs relying solely on the query or fixed predicate classifiers~\cite{krishna2017visual}. Moreover, most existing approaches extract and fuse embeddings from scene graphs, overlooking the fact that the structured textual representation is compatible with LLM-based reasoning on complex relational tasks~\cite{Struct-X,gu2025structext}. To address this, we propose a query-driven scene graph generation module that leverages VLMs to build scene graphs aligned with the query context. Our generated scene graphs include spatial relations, object-level descriptions, and inferred interactions, thereby bridging the gap between vision and language. By representing these graphs as structured text, we unlock the reasoning capabilities of LLMs, enabling more accurate grounding in complex visual scenes.

In this paper, we propose SGREC, a novel framework for zero-shot REC with query-driven scene graphs. Unlike existing methods that use CLIP or other pre-trained VLMs to align image regions with text tokens for target object localization, SGREC represents visual scenes using query-driven scene graphs that explicitly encode object relationships within the scene. First, SGREC identifies query-related objects by matching class labels with nouns, categories, and subjects. Next, it generates scene graphs incorporating the coordinates, captions, and interactions of the query-related objects, providing a detailed and structured description of the visual scene. Finally, the generated scene graphs along with the query are fed into LLMs to effectively infer the target object.

To evaluate the effectiveness of SGREC, we validate our framework using LLMs and VLMs and conduct extensive experiments on widely used REC benchmarks: RefCOCO, RefCOCO+, and RefCOCOg. The results show that SGREC achieves leading performance across these datasets, with a particularly notable improvement on the complex RefCOCOg, highlighting its strong visual scene understanding capabilities.

The main contributions of our work are as follows:
\begin{itemize}
	\item This paper proposes a novel framework for zero-shot referring expression comprehension, which integrates scene graphs and large language models (LLMs) to achieve a comprehensive understanding of visual scenes for accurate target object localization.
	\item We introduce a novel scene graph generation module that captures spatial information, object captions, and interactions in the image, providing a structured and detailed input for LLM-based inference.  
	\item Extensive experiments and ablation studies conducted on three widely used REC benchmarks demonstrate that our proposed method achieves leading performance and effectively validates its efficacy.
	\end{itemize}

\section{Related Work}

\subsection{Zero-shot Referring Expression Comprehension}
Zero-shot REC~\cite{subramanian2022reclip} focuses on transferring existing knowledge to the REC task without requiring task-specific training data, emphasizing the model's ability to generalize to new queries and objects.

Current zero-shot REC approaches typically employ VLMs to interpret queries and localize the corresponding target objects, with CLIP being the most widely adopted backbone. ReCLIP~\cite{subramanian2022reclip}, one of the earliest methods, extracts visual and textual embeddings via CLIP and computes similarity scores to identify the most relevant region, further introducing a spatial adjustment module to refine proposal distributions based on spatial relationships. Building upon this, RedCircle~\cite{shtedritski2023does} utilizes a red-circle visual prompt to enhance localization, while FGVP~\cite{yang2024fine} improves precision through diverse visual prompting strategies.
Considering that CLIP operates like a ``bag-of-words" model~\cite{yuksekgonul2022and} and struggles to reason about object relationships, recent zero-shot REC~\cite{han2024zero} aligns image regions with relation triplets extracted from the query. GroundVLP~\cite{shen2024groundvlp} further explores grounding via alternative VLMs such as VinVL~\cite{zhang2021vinvl} and ALBEF~\cite{li2021align}, both pre-trained on large-scale datasets containing annotations for objects, attributes, and relationships.
In addition, multimodal large language models such as KOSMOS-2~\cite{peng2023kosmos} and CoVLM~\cite{li2023covlm} have merged text spans and image regions within textual responses to infer referents directly. ViperGPT~\cite{suris2023vipergpt} and EAGR~\cite{bu2025error} adopt multi-stage pipelines that integrate several specialized models and extensive LLM-based code generation (e.g., GPT-3 Codex), whereas SoM~\cite{yang2023set} employs mark-based visual prompting for segmentation, which may introduce ambiguity or interference in crowded scenes due to overlapping or imprecise markings. In supervised settings, LLM-based frameworks such as FERRET~\cite{you2023ferret} require fine-tuning on large-scale dialogue data generated by GPT-4 to achieve robust comprehension.

In contrast, SGREC bridges the visual–textual semantic gap by generating query-conditioned scene graphs that explicitly model visual scenes for LLM inference. Instead of aligning image regions with language tokens, SGREC’s zero-shot scene graph provides a structured, supervisory-free representation, leveraging the LLM’s reasoning over text without the need for extensive fine-tuning or visual marking.

\subsection{Scene Graphs for Vision-language tasks}
Scene graphs have become a powerful tool in various vision-language tasks, including image captions~\cite{anderson2016spice}, object retrieval~\cite{johnson2015image}, 3D scene understanding~\cite{gu2024conceptgraphs}, and visual question answering~\cite{wang2023vqa}. They represent images as structured data, where nodes represent objects and edges describe their relationships. For example, ConceptGraphs~\cite{gu2024conceptgraphs} builds 3D scene graphs from RGB images by leveraging LLMs to infer spatial relationships as edges, while VQA-GNN~\cite{wang2023vqa} aligns scene graphs and concept graphs to transfer multi-modal knowledge. SGMN~\cite{yang2020graph} and LGRAN~\cite{wang2019neighbourhood} leverage pre-computed graphs generated by NLP parsers and require dedicated training for graph propagation and feature alignment between visual and linguistic features. Despite recent advancements in scene graphs for vision-language tasks, no existing REC methods have applied scene graphs to zero-shot REC tasks. The main challenges are as follows: 1) Detailed scene graph requirements. REC demands scene graphs capable of open-vocabulary relationship prediction and a comprehensive understanding of image content, which exceeds the capability of existing approaches. 2) Integration into answer prediction. Previous methods typically fuse node and edge embeddings in scene graphs for answer prediction, which complicates the generation of fine-grained representations necessary for accurately identifying objects and their relationships.

To overcome these challenges, we design a query-driven scene graph generation module that constructs scene graphs conditioned on the input query. Leveraging powerful VLMs, our approach captures spatial relations, object-level details, and implicit interactions to better align with the query’s intent and context.

\begin{figure*}
    \centering
    \includegraphics[scale=0.5]{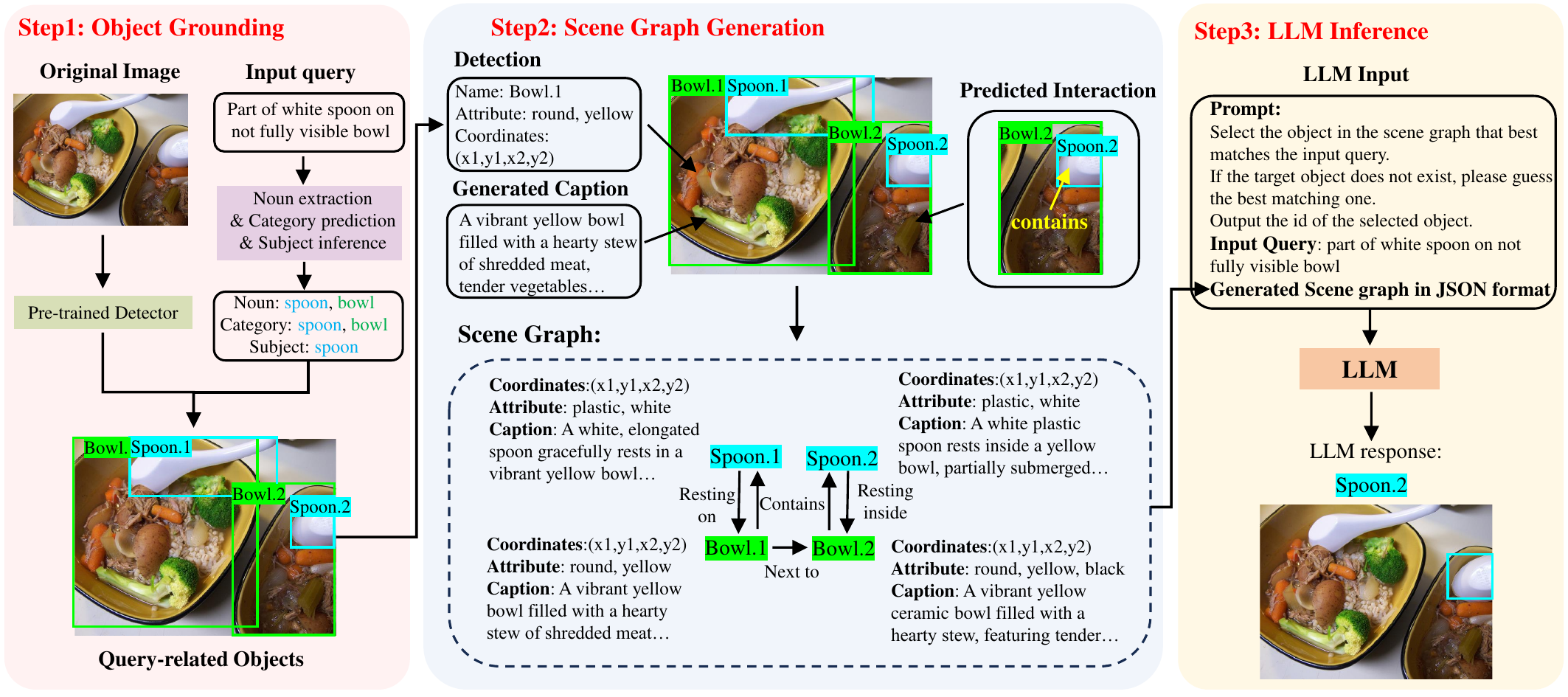}
 \caption{
 Pipeline of the proposed SGREC. In \textbf{Step 1}, SGREC begins by extracting nouns, predicting categories, and inferring subjects from the input query and the original image. It then identifies query-related objects by selecting those with matched labels. In \textbf{Step 2}, a scene graph is generated in three parts: class names and coordinates from the detector to encode spatial information, generated image captions for each object to describe their appearance, and predicted relation triplets to capture interactions between objects. Finally, in \textbf{Step 3}, SGREC analyzes the query and the generated scene graph to infer the index of the target object.}
 \label{fig_framework}
\end{figure*}

\section{The Proposed Method}
In this work, we introduce a novel framework, SGREC, for zero-shot REC that integrates scene graphs with interpretable inference using LLMs. An overview of our approach is illustrated in Figure~\ref{fig_framework}. 
In \textbf{Step 1}, we filter detected objects in the image by extracting nouns, categories, and subjects from the query, ensuring that only query-related visual objects are identified (Sec.~\ref{entity grounding}). In \textbf{Step 2}, we generate scene graphs to comprehensively represent the visual scene, incorporating spatial information, image captions, and object interactions (Sec.~\ref{Scene Graph Generation}). In \textbf{Step 3}, we leverage LLM to infer the target object based on the query and query-driven scene graph (Sec.~\ref{Context-aware Reasoning with LLMs}).

\subsection{Object Grounding}
\label{entity grounding}
Constructing scene graphs from an image begins with grounding query-related objects in the image. This process involves identifying relevant object labels and their corresponding bounding box coordinates. Object grounding comprises three key steps: noun extraction, category prediction, and subject inference. Finally, the query-related objects are selected with similar labels with nouns, categories, and subjects extracted from the queries.

Given one input image, we employ the VinVL detector~\cite{zhang2021vinvl} to detect all objects $O_{det}=\left\{\boldsymbol{o}_i\right\}_{i=1}^N$, where $N$ denotes the number of all objects. Each detected object $o_i$ is represented as $o_i=(l_i, p_i, a_i)$, consisting of its class label $l_i$, bounding box coordinate $p_i$, and attribute information $a_i$.

\noindent \textbf{Noun Extraction, Category Prediction \& Subject Inference} 
To retain query-related objects for scene graph construction, we extract nouns from the query, predict their corresponding categories, and infer the query’s subjects, which serve as the basis for object selection.

Existing methods typically rely on noun chunks~\cite{subramanian2022reclip} or combinations of nouns and adjectives~\cite{liu2023confidence} to identify objects, and then measure the feature similarities between nouns and detected class labels to select candidate objects. However, this approach can fail when there is a semantic gap between nouns and class labels, such as between ``mom" and ``person," potentially leading to missed objects. Moreover, some ambiguous queries may not include nouns (e.g., ``left thing'') that are specific enough to identify the target object, which inevitably leads to missing key objects. To address this, SGREC incorporates all extracted nouns along with their most semantically similar predicted category labels. Additionally, we leverage a VLM to infer subject labels, enriching the query-related object names.

Specifically, we employ SpaCy~\cite{honnibal2017spacy} to extract nouns and map them to category names defined in COCO~\cite{lin2014microsoft}. For subject inference as shown in Fig~\ref{hallu_sub}, we prompt LLaVA~\cite{liu2023llava} with both the image and the query using the instruction: \textit{\textbf{``Extract the subject of the query based on the image."}}. The model then produces a concise subject name (e.g., “table” for the ambiguous phrase “left thing”), which allows us to disambiguate query expressions and capture query-relevant subjects for constructing more comprehensive scene graphs. Finally, we collectively refer to the outputs of noun extraction, category prediction, and subject inference as the query-related object name set $O_{name}$, which consists of multiple object names associated with the query.

\begin{figure*}
    \centering
    \includegraphics[scale=0.36]{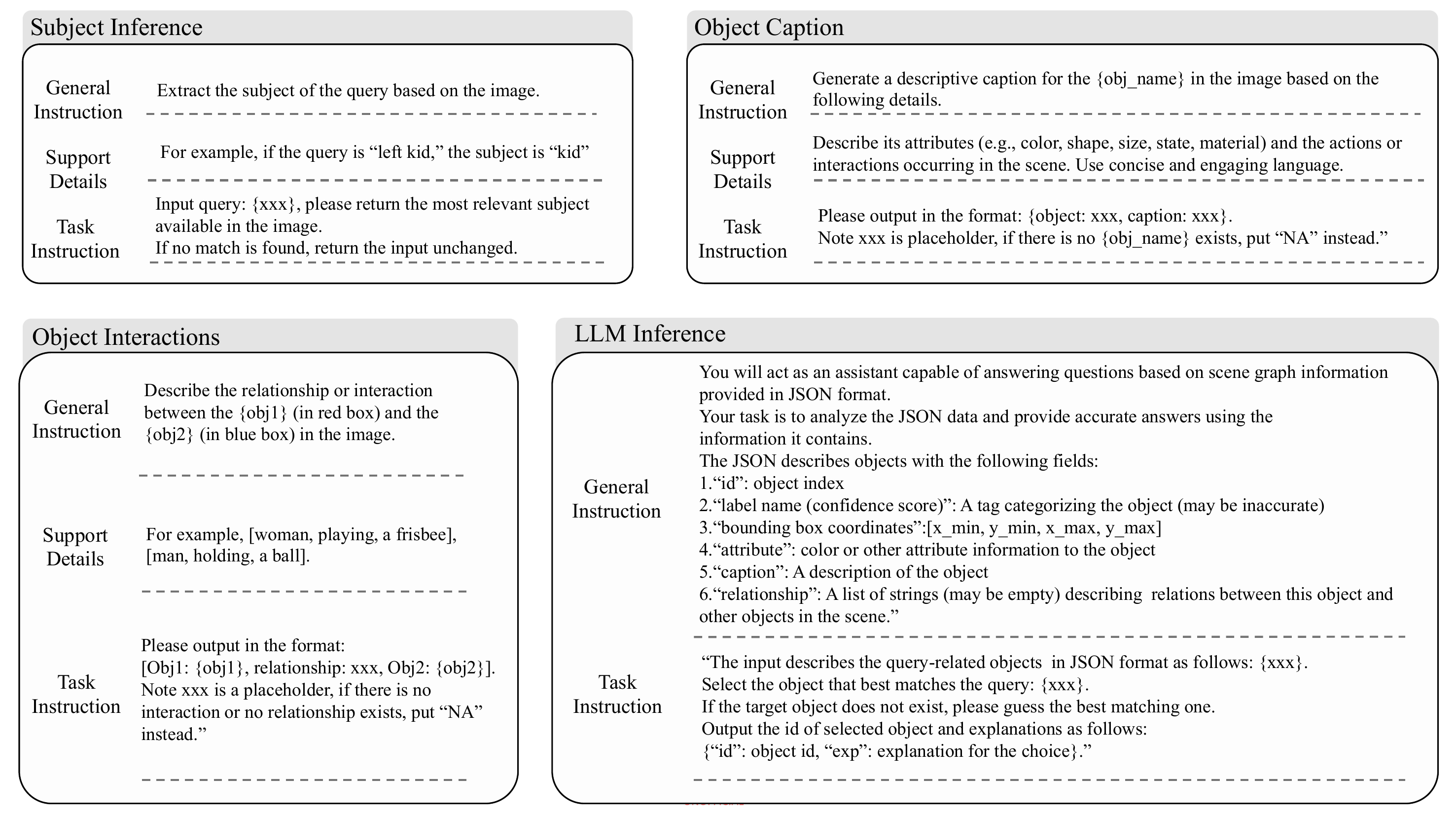}
 \caption{
 Detailed prompts used in SGREC, covering modules for subject inference, object caption generation, interaction extraction, and final LLM-based inference.}
 \label{prompts}
\end{figure*}

\begin{figure}[h]
\centering
\includegraphics[width=0.85\linewidth]{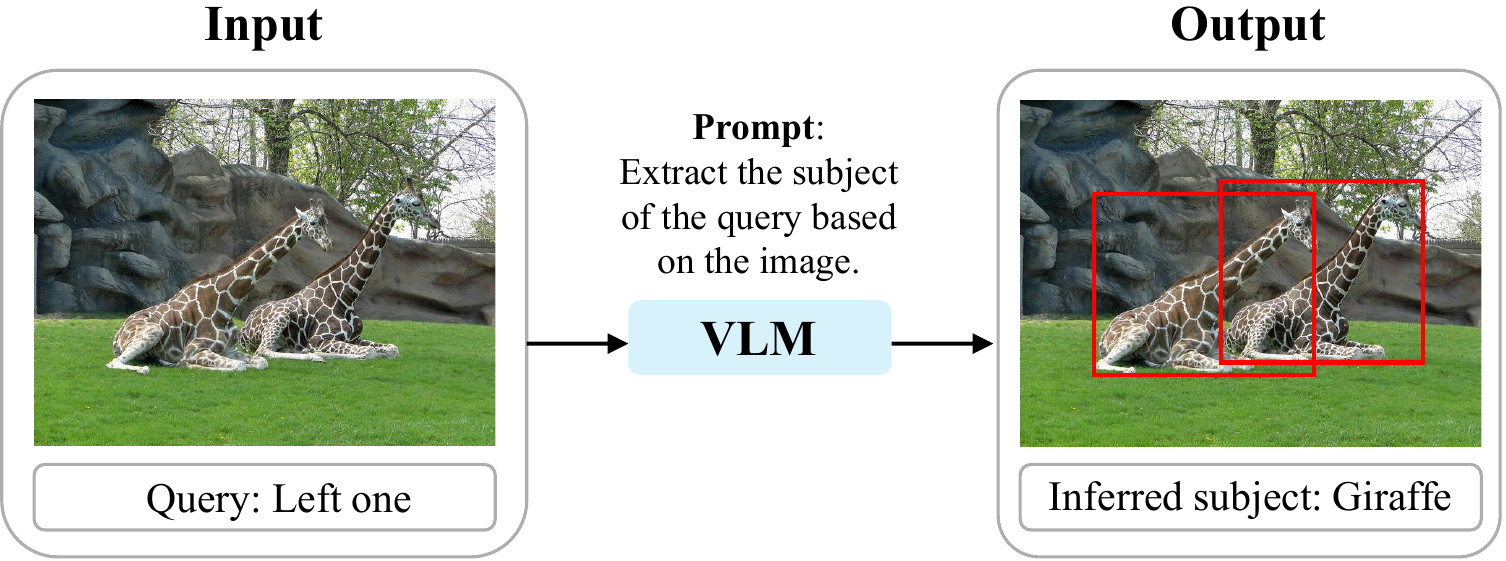}
\caption{Illustration of the subject inference, including its inputs and outputs.}
\label{hallu_sub}
\end{figure}

\noindent \textbf{Query-related Objects}
Given the query-related object names $O_{name}$, we select detected objects by matching their class labels, ensuring that as many query-related objects as possible are retained for subsequent scene graph generation. Each word is encoded with 300-dimensional word2vec embeddings~\cite{mikolov2013efficient}, and cosine similarity is computed between object names and class labels to preserve detected objects aligned with the query. Formally, the preserved query-related objects $O_{\text{sel}}$ are defined as:
\begin{equation}
O_{\text{sel}}=\Bigl\{\, o_i \in O_{\text{det}} \;\Big|\; 
\max_{n \in O_{\text{name}}}\,
\cos\!\bigl(\operatorname{emb}(l_i),\,\operatorname{emb}(n)\bigr)\ge \tau
\Bigr\}.
\end{equation}
where $\operatorname{emb}(\cdot)$ denotes the word2vec embedding function, $\cos(\cdot,\cdot)$ is the cosine similarity, and $\tau$ is the similarity threshold.

\subsection{Scene Graph Generation}
\label{Scene Graph Generation} 
The scene graph $SG$ is defined as $SG=(V,E)$, where $V=\{v_1, \dots, v_n\}$ denotes the set of query-related objects (the nodes) and $n$ is the total number of these objects. The edges $E$ represent the relationships between objects. Each node $v_i$ corresponds to an object and is represented as $v_i = (p_i, a_i, c_i))$, where $p_i$ denotes spatial information, $a_i$ denotes object attributes, and $c_i$ provides object captions. The edge $E_{i,j}$ captures the interactions between objects $v_i$ and $v_j$. This scene graph not only encodes the spatial locations and attributes of individual objects but also captures their interactions with other objects, providing a comprehensive representation of the entire scene.

\noindent \textbf{Spatial Information}
The spatial information $p_i$ is derived from the bounding box coordinates of each object. Previous methods~\cite{subramanian2022reclip,wang2024omni} rely on an additional spatial module to calculate relationships such as ``left" or ``bigger" by measuring distances between coordinates. These methods also require predefining the number of relationships, which often struggle with complex relationships like ``top right" or ``second left". Inspired by the numerical reasoning capabilities of LLMs~\cite{ahn2024large,yuan2025mme}, SGREC incorporates bounding box coordinates into the scene graphs, enabling LLMs to perform basic calculations and deduce spatial relationships between objects during the inference stage.

\noindent \textbf{Attribute Information}
The attribute information $a_i$ is also obtained from the detector. Similar to previous method~\cite{liu2023confidence}, we use a detector pre-trained on Visual Genome~\cite{krishna2017visual}, which outputs descriptive words related to color and state (e.g., ``yellow" and ``standing"). However, since this detector is limited to a closed set of attributes, it often fails to provide accurate descriptions of objects, necessitating the use of object captions to supplement attribute information for a more comprehensive representation.

\noindent \textbf{Object Caption} 
In this work, we introduce captions to supply richer contextual and relational cues that are not explicitly encoded in the attribute list. These captions convey additional cues such as human actions, numerical references, and textual elements present in the image, along with other fine-grained visual details that demand deeper scene understanding. This is particularly important for handling queries with highly specific descriptions, such as ``vase in a weird shape" or ``vase Figure 8", which require precise discrimination between objects exhibiting subtle differences in shape or style.

While certain weakly-supervised~\cite{liu2023confidence} or unsupervised methods~\cite{jiang2022pseudo} attempt to generate region captions to describe objects, these captions are often rule-based, combining phrases with nouns, colors, and spatial relationships, and lack detailed descriptions of image styles or shapes. To address this limitation, SGREC crops image regions using their bounding box coordinates and inputs each region into LLaVA with the prompt:``\textit{\textbf{Generate a descriptive caption for the \{obj\_name\} in the image. Describe its attributes, actions, or interactions occurring in the image.}}". This ensures that each region receives a concise and informative caption $c_i$ that accurately describes its content.

\begin{figure}
    \centering
    \includegraphics[scale=0.4]{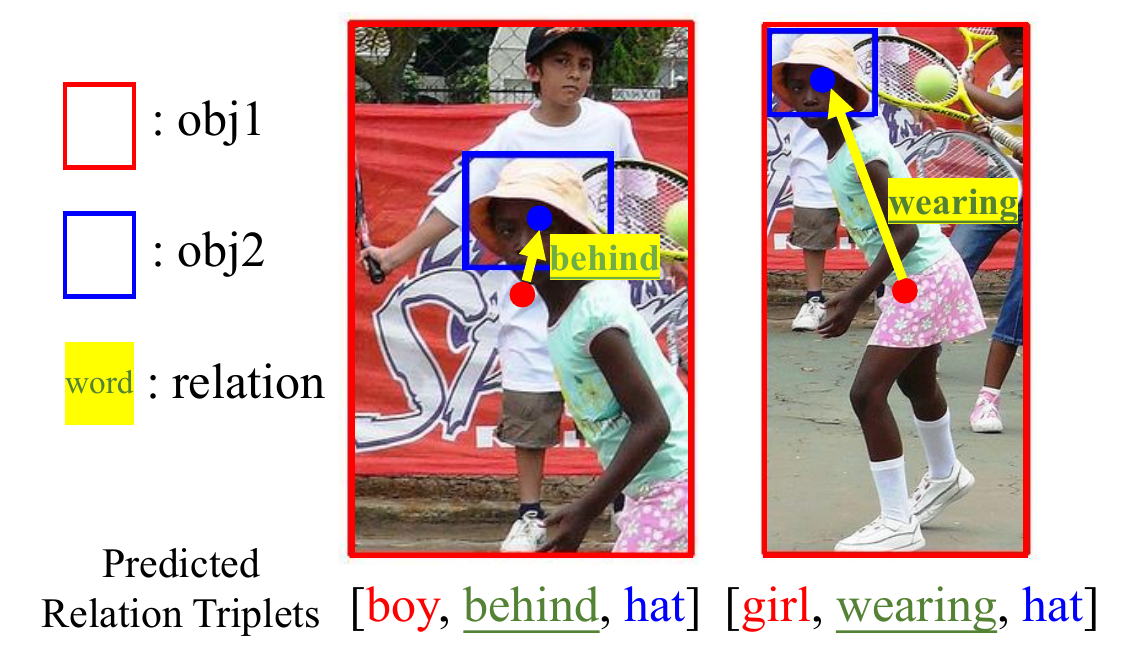}
 \caption{We highlight the obj1 and the obj2 using different colors to reduce confusion in scenarios where multiple similar objects appear.
 }
 \label{fig_interaction}
\end{figure}

\noindent \textbf{Interaction Construction} 
As shown in Fig.~\ref{fig_interaction}, we present object interactions $E_{i,j}$ in the scene by constructing relation triplets in the format of \{obj1:xxx, relation:xxx, obj2:xxx\}, where obj1 denotes $o_i$ and obj2 denotes $o_j$ sampled from query-related objects. Rather than relying on a predicate classifier restricted to a predefined set of predicates~\cite{zhang2021vinvl} or aligning images with relationships derived from the query~\cite{han2024zero}, we use LLaVA to predict these relationships directly from the image, offering greater flexibility and adaptability to new objects and predicates in real-world scenarios. To form each triplet, we first identify potential object pairs by calculating the ratio of the overlapping area between two objects relative to the smaller of those two objects. If this ratio exceeds a threshold $\theta$, we assume that a potential interaction may exist. We then prompt LLaVA to determine the specific relationship by instructing it to ``\textit{\textbf{Describe the relationship or interaction between the obj1 (in the red box) and the obj2 (in the blue box) in the image.}}". To reduce confusion when multiple similar objects, we visually highlight the obj1 and obj2 with red and boxes, respectively, providing a clear distinction.

\begin{table*}[t!]
\centering
\caption{Performance comparison of top-1 accuracy (\%) with zero-shot REC methods on RefCOCO, RefCOCO+, and RefCOCOg datasets. The best and second-best results are \textbf{boldfaced} and \underline{underlined}, respectively. ``Avg" denotes the average performance across all splits.}
\setlength{\tabcolsep}{1.7mm}{
\begin{tabular}{l|l|ccc|ccc|cc|c}
\hline
\multicolumn{1}{l|}{\multirow{2}{*}{\bf Method}} & \multirow{2}{*}{\bf Inference Model}  & \multicolumn{3}{c|}{RefCOCO}                     & \multicolumn{3}{c|}{RefCOCO+}                   & \multicolumn{2}{c|}{RefCOCOg}  & \multirow{2}{*}{\bf Avg} \\
\multicolumn{1}{c|}{}       &\multicolumn{1}{c|}{}                                       & \bf val            & \bf testA          & \bf testB          & \bf val            & \bf testA         & \bf testB          & \bf val          & \bf test         &                      \\ \hline
GPT-4V~\cite{achiam2023gpt}               & GPT-4V                                        & 25.48           & 26.22           & 24.39           & 10.59          & 18.23          & 8.87           & 14.26           & 15.42           & 17.93                \\
CPT~\cite{yao2024cpt}                & VinVL                                       & 32.20           & 36.10           & 30.30           & 31.90           & 35.20          & 28.80           & 36.70           & 36.50           & 33.46                \\
ReCLIP~\cite{subramanian2022reclip}    & CLIP ViT-B                                                & 45.78           & 46.10           & 47.07           & 47.87           & 50.10          & 45.10           & 59.33           & 59.01           & 50.05                \\
RedCircle~\cite{shtedritski2023does}       & CLIP ViT-L                                           & 49.80           & 58.60           & 39.90           & 55.30           & 63.90 & 45.40           & 59.40          & 58.90           & 53.90                \\
Pseudo-Q~\cite{jiang2022pseudo}   & TransVG & 56.02           & 58.25           & 54.13           & 38.88           & 45.06 & 32.13           & 46.25           & 47.44           & 47.27                \\
FGVP~\cite{yang2024fine}    &  SAM-ViT-H, CLIP ViT-L, CLIP RN50                                                    & 59.60  & 65.00  & 52.00  & 60.00  & 66.80 & 49.70  & 63.30           & 63.40           & 59.97       \\
CoVLM~\cite{li2023covlm}      & CLIP ViT-L, Pythia                                                 & 49.32          & 53.67          & 44.49          & 48.87          & 52.51         & 44.71          & 61.23          & 62.33          & 52.14                \\
KOSMOS-2~\cite{peng2023kosmos}       &  MAGNETO                                            & 52.32          & 57.42          & 47.26          & 45.48          & 50.73         & 42.24          & 60.57          & 61.65          & 52.21                \\
ViperGPT~\cite{suris2023vipergpt}       &  GLIP, X-VLM, MiDaS, BLIP2, GPT-3                                            & -          & 62.80          & 51.20          & -          & 61.00         & 47.20          & -          & 50.40          & 54.52                \\
GroundVLP(ALBEF)~\cite{shen2024groundvlp}      & ALBEF                                             & 52.58          & 61.30         & 43.53          & 56.38          & 64.77         & 47.43         & 64.30          & 63.54          & 56.73                \\
GroundVLP(VinVL)~\cite{shen2024groundvlp}      & VinVL                                             & 59.05          & 69.21         & 48.71          & \textbf{61.80}          & \textbf{70.56}         & 50.97         & \underline{69.08}          & 68.98          & 62.29                \\
Pink~\cite{xuan2024pink}   & CLIP ViT-L, Vicuna-7B                            & 54.10          & 61.20           & 44.20          & 43.90          & 50.70         & 35.00          & 59.10           & 60.10          & 51.04                \\
ZeroshotREC~\cite{han2024zero}   & CLIP ViT-B                                             & 48.24          & 48.40           & 49.15          & 45.64          & 47.59         & 42.79          & 57.6           & 56.64          & 49.51                \\
ZeroshotREC(VRCLIP)~\cite{han2024zero}   & CLIP ViT-B                                             & 60.62          &  66.52           & 54.86          & 55.52          & 62.56         & 45.69          & 59.87           & 59.90          & 58.19                \\
MCCE-REC~\cite{qiu2024mcce}      & CLIP ViT-B, LLaVA-13B                                             & \underline{60.78}         & 68.99         & 53.33          & \underline{59.03}          & \underline{67.81}         & 48.13         & 62.56          & 62.30          & 60.37  \\     EAGR~\cite{bu2025error}   & GLIP, X-VLM, MiDaS, BLIP2, GPT-3.5                    & -          & \underline{71.00}           & \underline{63.80}          & -          & 64.00         & \textbf{53.60}          & -           & \underline{71.40}          & \underline{64.76}                \\          
\hline
\textbf{SGREC}                   & LLaVA-7B, Qwen-72B                        & \textbf{66.78}             & \textbf{71.38}                & \textbf{64.14}               & 57.17               & 61.49          & \underline{53.43}  & \textbf{73.28}           & \textbf{72.97}                           & \textbf{65.08}        \\ \hline
\end{tabular}}
\label{main_table}
\end{table*}

\begin{table*}[]
\centering
\caption{Performance comparison under zero-shot, weakly supervised, and fully supervised settings on RefCOCO/+/g datasets. The best results are \textbf{boldfaced}. ``Avg" denotes the average performance across all splits.}
\label{full}
\setlength{\tabcolsep}{2.6mm}{
\begin{tabular}{l|c|ccc|ccc|cc|c}
\hline
\multirow{2}{*}{Model} & \multirow{2}{*}{Setting} & \multicolumn{3}{c|}{RefCOCO}                     & \multicolumn{3}{c|}{RefCOCO+}                    & \multicolumn{2}{c|}{RefCOCOg}   & \multirow{2}{*}{\textbf{Avg}} \\
                       &                          & val            & testA          & testB          & val            & testA          & testB          & val            & test           &                               \\ \hline
LGRAN~\cite{wang2019neighbourhood}                & \multirow{4}{*}{Fully-supervised}                     & 82.00         & 81.20          & 84.00    & 66.60         & 67.60           & 65.50         & 75.40          & 74.70           & 74.62  \\
SGMN~\cite{yang2020graph}                &                      & -          & 86.67          & 85.36 
& -         & 78.66           & 69.77         & -          & 81.42           & 80.37  \\
Ferret-13B~\cite{you2023ferret}                &                      & 89.48          & 92.41          & 84.36 
& 82.81         & 88.14           & 75.17         & 85.83          & 86.34           & 85.57  \\
Grounding-DINO~\cite{liu2024grounding}         &                      & 90.56          & 93.19          & 88.24          & 82.75          & 88.95          & 75.92          & 86.13          & 87.02          & 86.60                         \\ \hline
Cycle-Free~\cite{sun2021cycle}             & \multirow{4}{*}{Weakly-supervised}    & 39.58          & 41.46          & 37.96          & 39.19          & 39.63          & 37.53          & -              & -              & 39.23                         \\
CPL~\cite{liu2023confidence}                    &                          & 70.67          & 74.58          & 67.19          & 51.81          & 58.34          & 46.17          & 60.21          & 60.12          & 61.14                         \\
APL~\cite{luo2024apl}                    &                          & 64.51          & 61.91          & 63.57          & 42.70          & 42.84          & 39.80          & 50.22          & -              & 52.22                         \\
AlignCAT~\cite{wang2025aligncat}               &                          & 69.03          & 70.27          & 66.59          & 47.16          & 52.22          & 41.91          & 54.72          & -              & 57.41                         \\ \hline
Grounding-DINO~\cite{liu2024grounding}         & \multirow{2}{*}{Zero-shot}    & 50.41          & 57.24          & 43.21          & 51.40          & 57.59          & 45.81          & 67.46          & 67.13          & 55.03                         \\
SGREC (Ours)           &                          & \textbf{66.78} & \textbf{71.38} & \textbf{64.14} & \textbf{57.17} & \textbf{61.49} & \textbf{53.43} & \textbf{73.28} & \textbf{72.97} & \textbf{65.08}                         \\ \hline
\end{tabular}}
\end{table*}

\subsection{LLM Inference}
\label{Context-aware Reasoning with LLMs}
The query and scene graph are input to an LLM, along with an instruction: \textit{\textbf{Select the object in the scene graph that best matches the input query.”}}. The LLM then returns the index of the target object with the explanation for its choice. The selected object's index is then used to retrieve its bounding box coordinates in the query-related objects.

\section{Experiments}

\subsection{Datasets}
We conduct experiments on RefCOCO/+/g datasets collected from MS-COCO~\cite{lin2014microsoft}. The dataset splits follow previous works~\cite{ subramanian2022reclip,han2024zero}.

\textbf{RefCOCO}~\cite{yu2016modeling} consists of 19,994 images and 142,210 referring expressions, emphasizing spatial-related descriptions.
\textbf{RefCOCO+}~\cite{yu2016modeling} contains 19,992 images and 141,564 expressions, focusing on appearance-related descriptions. Both RefCOCO and RefCOCO+ are divided into three splits, where testA consists of persons as the target objects and testB covers other object types. \textbf{RefCOCOg}~\cite{mao2016generation} has 25,799 images and 95,010 expressions, which are generally more detailed and longer.

\subsection{Evaluation Metrics}
Following existing methods~\cite{subramanian2022reclip,han2024zero}, we adopt top-1 accuracy as our performance metric. The performance is measured by computing the Intersection over Union (IoU) between the predicted and ground-truth bounding boxes; if the IoU exceeds 0.5, it is considered a correct prediction.

\subsection{Implementation Details}
We employ the same detector used in CPL~\cite{liu2023confidence}, which generates class labels, bounding box coordinates, and attribute information for all region proposals, taking all these proposals as input. For the noun extraction module, we use SpaCy's\footnote{\url{https://spacy.io/}}~\cite{honnibal2017spacy} Part-of-Speech (POS) tagger to identify nouns (i.e., terms tagged as ``NOUN", ``PROPN", and ``PRON") and utilize a 300-dimensional word2vec embedding\footnote{\url{https://code.google.com/archive/p/word2vec/}}~\cite{mikolov2013efficient} for each word to calculate similarity between nouns and categories. We use LLaVA~\cite{liu2023llava} (LLaVA-onevision-qwen2-ov-chat) as the VLM and several LLMs: LLaMA~\cite{touvron2023llama} (Llama-3.1-Instruct) and Qwen~\cite{yang2024qwen2} (Qwen2.5-Instruct-GPTQ-Int4). The sampling parameters are set to their default values. For Qwen2.5, the temperature is 0.7 and top\_p is 0.8, while for LLaMA, the temperature is 0.6 and top\_p is 0.9. The threshold $\tau$ for selecting detected objects is set to 0.5 in Table~\ref{tau}.
The threshold $\theta$ for identifying potential interactions between objects is set to 0.2 in Table~\ref{threshold}. Detailed prompts are provided in Figure~\ref{prompts}. All experiments are conducted on one NVIDIA H100 GPU.

\subsection{Main results}
\noindent \textbf{Comparison with State-of-the-art Zero-shot Methods}
As illustrated in Table~\ref{main_table}, SGREC achieves the highest average top-1 accuracy. On RefCOCO, which mainly focuses on spatial-related queries, SGREC surpasses ZeroshotREC(VRCLIP) by \textbf{4.86\%–9.28\%}. While ZeroshotREC(VRCLIP) is fine-tuned on data tailored for REC tasks, SGREC achieves superior performance without any fine-tuning. On RefCOCOg, which involves more complex and detailed queries, SGREC outperforms MCCE-REC by over \textbf{10\%}. In addition, on the testB split of RefCOCO+, SGREC outperforms existing methods by over \textbf{5\%}, suggesting that SGREC can effectively describe objects. These results highlight SGREC’s effectiveness in spatial localization, object descriptions, and robust modeling of object relationships, facilitated by the comprehensive representation of the visual scene through its generated scene graphs.

Furthermore, we observe that increasing model size alone does not guarantee higher accuracy. Methods such as ViperGPT~\cite{suris2023vipergpt} and EAGR~\cite{bu2025error}, which rely on much larger GPT models, are outperformed by SGREC-Qwen-72B on both RefCOCO and RefCOCOg. These results indicate that our improvement primarily stems from the structured relational modeling introduced by the scene graph, rather than from scaling up the LLM.

\begin{table*}[]
\centering
\caption{Influence of different inference models on RefCOCO, RefCOCO+, and RefCOCOg datasets. The best and second-best results are \textbf{boldfaced} and \underline{underlined}, respectively. ``Avg" denotes the average performance across all eight splits.}
\setlength{\tabcolsep}{3.5mm}{
\begin{tabular}{l|l|ccc|ccc|cc|c}
\hline
\multicolumn{1}{c|}{\multirow{2}{*}{\bf Method}}  & \multicolumn{1}{c|}{\multirow{2}{*}{\bf Inference}} & \multicolumn{3}{c|}{RefCOCO}                     & \multicolumn{3}{c|}{RefCOCO+}                   & \multicolumn{2}{c|}{RefCOCOg}  & \multirow{2}{*}{\bf Avg} \\
\multicolumn{1}{c|}{}          &\multicolumn{1}{c|}{}              & \bf val            & \bf testA          & \bf testB          & \bf val            & \bf testA         & \bf testB          & \bf val          & \bf test         &                      \\ \hline

\multirow{6}{*}{SGREC}  
& LLaMA-8B~\cite{touvron2023llama}                                               & 53.41 & 57.58  & 48.95  & 48.74 & 52.69       & 44.43 & 59.72  & 59.13  & 53.08       \\ 
   & LLaMA-70B~\cite{touvron2023llama}                                                & 64.04 & 69.26 & 59.82 & 55.85 & 60.93         & 51.16 & 71.28 & 71.23 & 62.95       \\ 
   & LLaVA-7B~\cite{liu2023llava}                                              & 45.58 & 50.72 & 42.32  & 45.37 & 48.67         & 42.58 & 56.64 & 56.38  & 48.53     \\ 
& LLaVA-72B~\cite{liu2023llava}                                            & \underline{65.86} & \underline{70.28}  & \underline{61.76}  & \textbf{57.53}  & \underline{61.35}         & \textbf{53.77} & \underline{73.10} & \underline{72.86}  & \underline{64.56}        \\ 
     & Qwen-7B~\cite{yang2024qwen2}                                              & 54.15 & 58.26  & 49.40 & 49.54 & 53.53         & 46.10 & 64.06 & 64.11 & 54.89       \\ 
  & Qwen-72B~\cite{yang2024qwen2}                                    & \textbf{66.78}             & \textbf{71.38}                &\textbf{64.14}               & \underline{57.17}               & \textbf{61.49}          & \underline{53.43}  & \textbf{73.28}           & \textbf{72.97}                            & \textbf{65.08}        \\ \hline
\end{tabular}}
\label{diff_llms}
\end{table*}

\noindent \textbf{Comparison with Fully and Weakly-Supervised Methods}
To further validate the effectiveness of SGREC in modeling visual relationships, we conduct comparisons against recent fully and weakly-supervised approaches. As reported in Table~\ref{full}, we benchmark SGREC against Grounding-DINO~\cite{liu2024grounding}, a state-of-the-art VLM detector tailored for this task that predicts bounding boxes conditioned on text queries, with both fully-supervised and zero-shot variants available. While Grounding-DINO achieves competitive results when trained on RefCOCO/+/g, its performance degrades significantly under zero-shot conditions. In contrast, SGREC consistently outperforms Grounding-DINO by a margin of 3–20\%, underscoring the effectiveness of scene graphs for representing image content in zero-shot scenarios. Moreover, SGREC achieves performance comparable to the fully supervised graph-based method LGRAN~\cite{wang2019neighbourhood} on the RefCOCOg, confirming its effectiveness in modeling complex visual relationships.

We further compare SGREC with weakly-supervised baselines. Results show that SGREC achieves notable improvements on RefCOCO+ and RefCOCOg, demonstrating the strong capability of scene graphs to model complex visual scenes and relationships even without task-specific training. In particular, SGREC surpasses CPL by more than \textbf{13\%} on RefCOCOg, highlighting its robustness in handling challenging queries that involve fine-grained relational reasoning.

In summary, these comparisons indicate that zero-shot REC remains a highly challenging setting without access to training data. Nevertheless, SGREC establishes a substantial advantage over weakly-supervised baselines, thereby validating the importance of scene graphs in enabling interpretable and effective zero-shot reasoning.

\noindent \textbf{Influence of Inference Models}
To assess the compatibility of scene graphs with different large language models (LLMs), we experiment with various models, including LLaMA, LLaVA, and Qwen. Since LLaVA is a multimodal model, we disable image input by setting it to None, using only scene graphs during inference to isolate its language reasoning ability.
As shown in Table~\ref{diff_llms}, larger models consistently achieve better performance across datasets, due to their stronger language comprehension and reasoning capabilities. These models excel at interpreting and inferring relationships between objects and queries, enhancing the effectiveness of scene graphs in representing visual scenes. This demonstrates that larger LLMs are more adept at processing complex queries and intricate visual contexts. 

Additionally, despite using smaller models, \textbf{SGREC-Qwen-7B} achieves over 5\% average improvement compared with ZeroshotREC (which relies on ChatGPT) and surpasses Pink (Vicuna-7B) by 3.85\% on average. On the challenging RefCOCOg benchmark, SGREC-Qwen-7B ranks 3rd on val (\textbf{1.5\%} above MCCE-REC) and 4th on test (\textbf{1.81\%} above MCCE-REC), showing that structured reasoning brings consistent gains even with compact LLMs.

\subsection{Analysis and Ablation Study}

\noindent \textbf{Ablation Study of Detected Objects} 
To evaluate the impact of different types of objects on our framework (Step 1), we conduct experiments that incorporate noun-based, category-based, and subject-based objects, as presented in Table~\ref{proposals}. These objects are critical for constructing scene graphs by determining the nodes. Our analysis reveals that directly using noun-based objects enables the localization of a majority of targets. By incorporating predicted category labels, we expand the semantic scope of nouns, resulting in a substantial performance improvement of 3.41\%–4.57\% on RefCOCO/+/g. Finally, including subject-based objects addresses ambiguities in expressions, further improving accuracy by 1.61\%–4.1\%.

\begin{table}
\centering
\caption{Ablation Study of Detected Objects on \textbf{val} split on RefCOCO/+/g. \textbf{Cat}: Predicted Category. \textbf{Sub}: Inferred Subject.}
\label{proposals}
\begin{tabular}{ccc|ccc}
\hline
\multicolumn{3}{c|}{Detected objects} & \multirow{2}{*}{RefCOCO} & \multirow{2}{*}{RefCOCO+} & \multirow{2}{*}{RefCOCOg} \\
Noun  & Cat & Sub &                           &                           &                          \\ \hline
$\checkmark$         &        &             & 58.11                         &   49.00                        & 68.26                         \\
$\checkmark$         & $\checkmark$        &             &  62.68                        &  53.28                         &  71.67                        \\
$\checkmark$         & $\checkmark$        & $\checkmark$           & \textbf{66.78}        & \textbf{57.17}          & \textbf{73.28}                          \\ \hline
\end{tabular}
\vspace{-5pt}
\end{table}

\begin{table}[t]
\centering
\caption{Ablation Study of different information contained in generated scene graphs on \textbf{val} split of RefCOCO/+/g. \textbf{Det}: Detected objects along with their coordinates and attribute information. \textbf{Cap}: object captions. \textbf{Inter}: semantic interactions.}
\begin{tabular}{ccc|ccc}
\hline
\multicolumn{3}{c|}{Scene Graph} & \multirow{2}{*}{RefCOCO} & \multirow{2}{*}{RefCOCO+} & \multirow{2}{*}{RefCOCOg} \\
Det  &  Cap &  Inter &                           &                           &                          \\ \hline
$\checkmark$         &        &             &  62.27                        & 47.68                          & 63.95                         \\
$\checkmark$         & $\checkmark$        &             &   65.82                       &  54.60                         &  70.30                        \\
$\checkmark$         & $\checkmark$        & $\checkmark$           & \textbf{66.78}        & \textbf{57.17}          & \textbf{73.28}                          \\ \hline
\end{tabular}
    \label{modules}
    \vspace{-5pt}
\end{table}

\begin{table}[t]
\centering
\caption{Influence of $\tau$ that retain detected objects on the \textbf{val} split of RefCOCO/+/g.}
\label{tau}
\setlength{\tabcolsep}{3.5mm}{
\begin{tabular}{c|ccc}
\hline
                        \bf $\tau $ & RefCOCO & RefCOCO+ & RefCOCOg \\ \hline
                           0.3      & 66.64        & 56.81          & 73.03         \\
                           0.4      & 66.73        & 57.01          & 73.14         \\
                           0.5      & \textbf{66.78}        & \textbf{57.17}          & \textbf{73.28}         \\
                           0.6      & 66.37         & 55.98         & 72.06         \\ 
                           0.7      & 65.19         & 55.43        & 71.52         \\\hline
\end{tabular}}
\end{table}

\noindent \textbf{Ablation Study of Scene Graphs} We evaluate the performance of SGREC with different types of information included in the scene graphs (Step 2) in Table~\ref{modules}. Spatial and attribute information, directly obtained from the detector, provides a basic representation of visual scenes using box coordinates and attribute words. By incorporating image captions, SGREC greatly improves its performance, as the captions provide more comprehensive descriptions of objects. Especially for RefCOCO+, which focuses on appearance-related descriptions, where captions boost performance by 6.92\%. Finally, modeling object interactions further enhances SGREC’s performance, achieving the best results. RefCOCOg, with its longer and more detailed expressions requiring a nuanced understanding of object relationships, can boost performance by 2.98\%. RefCOCO, which emphasizes spatial-related descriptions, modeling object interactions provides a smaller improvement of 0.96\%.

\begin{figure*}[t!]
    \centering
    \includegraphics[scale=0.19]{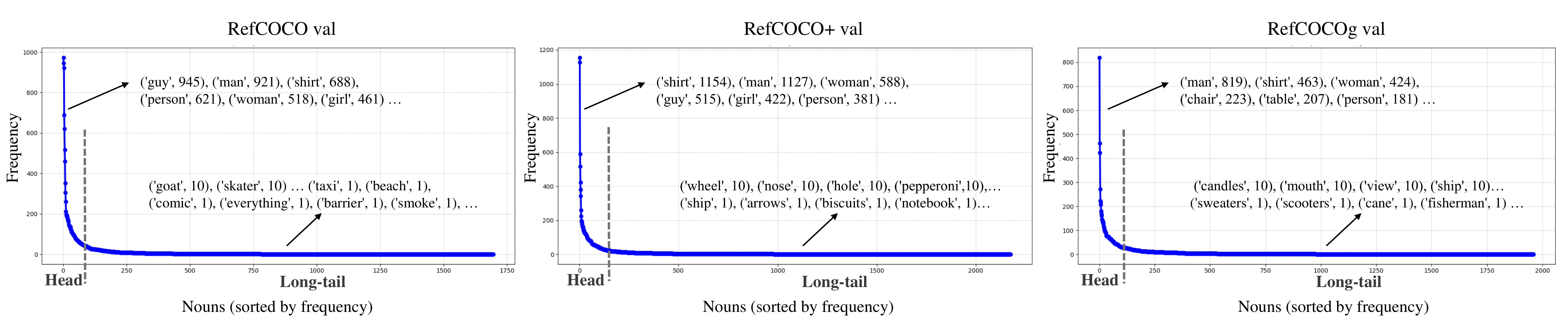}
 \caption{The frequency of noun occurrences in queries across different datasets. (`name', number) indicates how many times the corresponding noun appears.
 }
 \label{long-tail}
\end{figure*}

\begin{table}
\centering
\caption{Influence of $\theta$ that identify possible interactions on the \textbf{val} split of RefCOCO/+/g.}
\label{threshold}
\setlength{\tabcolsep}{3.5mm}{
\begin{tabular}{c|ccc}
\hline
                        \bf $\theta$ & RefCOCO & RefCOCO+ & RefCOCOg \\ \hline
 
                           0.1      & 66.62        & 56.39          & 72.94         \\
                           0.2      & \textbf{66.78}        & \textbf{57.17}          & \textbf{73.28}         \\
                           0.3      & 66.59         & 56.38         & 72.32         \\ 
                           0.4      & 66.46         & 56.16        & 72.27         \\ 
                           0.5      & 66.25         & 55.84        & 71.86                  \\\hline
\end{tabular}}
\end{table}

\noindent \textbf{Influence of Threshold $\tau$ on Retaining Detected Objects}
To ensure that only query-related detected objects are preserved for scene graph construction, we apply a threshold $\tau$ that filters out unrelated ones by computing the cosine similarity between candidate object names and object class labels. As shown in Table~\ref{tau}, when $\tau$ is set within the range [0.3, 0.5], the overall performance remains relatively stable. However, smaller $\tau$ values allow more irrelevant objects to pass through, thereby increasing inference complexity. Conversely, when $\tau$ is raised to [0.6, 0.7], the stricter filtering also eliminates some relevant objects, leading to a drop in performance.

\noindent \textbf{Influence of Threshold $\theta$ to Identify Possible Interactions}
Although SGREC identifies query-related objects, predicting interactions between any two objects is often unnecessary. For objects that are far apart, spatial relationships can be directly inferred from their coordinates. To filter object pairs and identify potential interactions, we introduce a threshold $\theta$. As shown in Table~\ref{threshold}, within the range of [0.1, 0.3], our method maintains comparable performance, demonstrating its robustness across a wide range of thresholds. When the threshold exceeds 0.3, we start to observe a slight degradation in performance. The main reason is higher thresholds often exclude many relevant object pairs, while lower thresholds tend to include loosely related or irrelevant ones, such as objects with minimal spatial overlap, which introduces noise and increases computational overhead.

\begin{table}[t]
\vspace{-5pt}
\setlength{\tabcolsep}{2mm}{
\centering
\caption{Model performance comparisons between high-frequency and low-frequency nouns on the val split of RefCOCO/+/g. Acc denotes accuracy.}
\begin{tabular}{c|cc|cc|cc}
\hline
\multirow{2}{*}{Freq of Nouns} & \multicolumn{2}{c|}{RefCOCO} & \multicolumn{2}{c|}{RefCOCO+} & \multicolumn{2}{c}{RefCOCOg} \\
                               & Queries         & Acc        & Queries         & Acc        & Queries        & Acc        \\ \hline
\textgreater{}200              & 5255                & 70.50           & 4223                & 61.61           & 2043               & 77.43           \\
100$\sim$200                   & 2551                &  66.13          & 2560                & 57.34           & 1163               & 72.92           \\
\textless{}=100                & 3028                &  60.87          & 3975                & 52.33           & 1690               & 68.52           \\ \hline
\end{tabular}}
    \label{freq}
    \vspace{-5pt}
\end{table}

\begin{table}[t]
\centering 
\caption{Accuracy across different decoding parameter settings on the \textbf{val} split of RefCOCOg.}
\begin{tabular}{l|cc|c}
\hline
        & Temperature & Top\_p & RefCOCOg val \\ \hline
Default & 0.7         & 0.8    & \textbf{73.28}            \\
Exp1    & 0.7         & 1.0    & 72.92            \\
Exp2    & 0.3         & 0.8    & 73.02            \\
Exp2    & 0.3         & 1.0    & 73.24            \\\hline
\end{tabular}
\label{parameter}
\end{table}

\noindent \textbf{Evaluating Long-Tail Generalization}
To investigate the model’s performance across varying query-object frequencies, we first analyzed the frequency of nouns in queries from each dataset's validation sets, using SpaCy for noun extraction. As shown in Figure~\ref{long-tail}, the distribution presents a long-tail pattern, with many low-frequency nouns. Furthermore, we observe that these datasets share similar high-frequency person-related nouns, like ``guy", ``person", and ``man". To examine whether our model exhibits bias toward frequently occurring nouns, we divided the nouns into three groups: high, mid, and low frequency. We then evaluate the model performance on each group separately as shown in Table~\ref{freq}. The results indicate that the model handles low-frequency categories well, demonstrating its ability to generalize to less common nouns.

\noindent \textbf{Consistency of Model Predictions}
We adopt the default decoding parameters for LLM inference in our main experiments. To assess the consistency of model predictions, we further test different temperatures and top-p values to analyze their effect on sampling stability. The results in Table~\ref{parameter} show that the model generates stable and consistent outputs across repeated runs, indicating strong robustness to prompt and sampling variations. 
Specifically, we vary the decoding temperature from 0.0 to 0.7 and top-p from 0.8 to 1.0. Lower temperatures lead to more deterministic reasoning, while higher ones increase output diversity. Smaller top-p values constrain sampling to high-probability tokens, producing more stable predictions. As shown in Table~\ref{parameter}, these variations cause only marginal performance differences, demonstrating that our framework remains robust under different decoding settings. Therefore, we use the default parameters (temperature = 0.7, top-p = 0.8) for all other experiments.

\begin{figure}[t]
\centering
\includegraphics[width=0.6\linewidth]{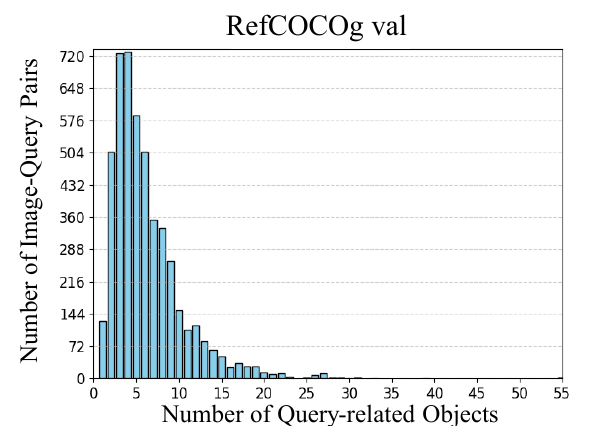}
\caption{Distribution of detected query-related objects across image-query pairs on the val split of RefCOCOg.}
\label{num_det}
\end{figure}

\noindent \textbf{Robustness under Dense Scenes}
We assess the model’s robustness under dense scenes, as presented in Figure~\ref{num_det} and Table~\ref{detobj}. As shown in Figure~\ref{num_det}, most images contain between 5 and 15 detected objects, while samples with more than 20 objects decrease sharply, though a few cases reach up to 50–55 detected objects. Based on this distribution, we categorize image–query pairs into five density levels according to the number of detected objects ([0, 5), [5, 10), [10, 15), [15, 20), and $\geq$20). 
Table~\ref{detobj} shows a gradual accuracy decline as the number of detected objects increases, reflecting stronger interference in denser scenes. Nevertheless, the model maintains stable performance within the common 0–20 object range, which covers most samples. Even under extremely dense conditions ($\geq$20 objects), it still achieves 77.97\% accuracy on RefCOCOg, surpassing prior zero-shot methods evaluated on the full validation set. These results highlight the model’s robustness and its strong ability to handle dense, cluttered visual scenes.

\begin{table}[t]
    \centering
    \caption{Model performance comparison between sparse and dense scenes on the \textbf{val} split of RefCOCOg. Acc denotes accuracy. Samples denotes the number of image-query pairs.}
    \begin{tabular}{c|cc}
    \hline
    \multirow{2}{*}{Num of Objects}  & \multicolumn{2}{c}{RefCOCOg} \\
                                    & Samples        & Acc        \\ \hline
    $[0,5)$                           & 2089               & 74.63           \\
    $[5,10)$                         & 2045               & 73.49           \\
    $[10,15)$                         & 527               & 70.78           \\
    $[15,20)$                         & 162               & 59.87           \\
    $\geq$ 20                          & 59               & 77.97           \\ \hline
    \end{tabular}
        \label{detobj}
    \end{table}

\begin{table}[t]
\caption{Examples of three kinds of input forms and corresponding results on the RefCOCOg val.}
\label{examples}
\centering
\setlength{\tabcolsep}{1.5mm}{
\begin{tabular}{l|l|c}
\hline
\multicolumn{1}{c|}{Type}   &\multicolumn{1}{c|}{Example} & \multicolumn{1}{c}{RefCOCOg val}                                                                                                                           \\ \hline
Natural language       & `a man wearing a red...' &  62.01                                                                                           \\
Structured text       & Object 1. label: man, attribute:...     & 72.79 \\
JSON               & \{`id':1, `label':[man], `attribute':[...] \}          &  73.28                                                            \\ \hline
\end{tabular}}
\end{table}

\noindent \textbf{Input format of Scene Graphs}
In our experiments, the generated scene graph is serialized into a \textbf{plain-text JSON string} and directly fed into the LLM prompt. To illustrate the difference between possible input forms, we provide examples and corresponding results in Table~\ref{examples}. As shown, the natural-language prompt is expressed entirely in free-form language, the structured text presents information through fixed slots and labels without explicit hierarchy, while the JSON format encodes the same information with explicit key–value pairs with clear structural relations.
In our ablation, the natural-language setting corresponds to using only the generated captions, whereas the structured-text setting flattens the JSON representation by removing structural symbols but retaining field labels. The results indicate that preserving structural cues in textual form improves reasoning stability and grounding accuracy.

\noindent \textbf{Inference Time}
We measured the average runtime per image-query pair on
a single H100 GPU to assess inference efficiency. Subject
inference takes around 0.16 seconds, scene graph generation
takes 1.11 seconds (split between 0.59 seconds for captioning
and 0.52 seconds for relation inference), and the final LLM
reasoning step takes 7.67 seconds using Qwen-72B.

\begin{figure*}[t!]
    \centering
    \includegraphics[scale=0.42]{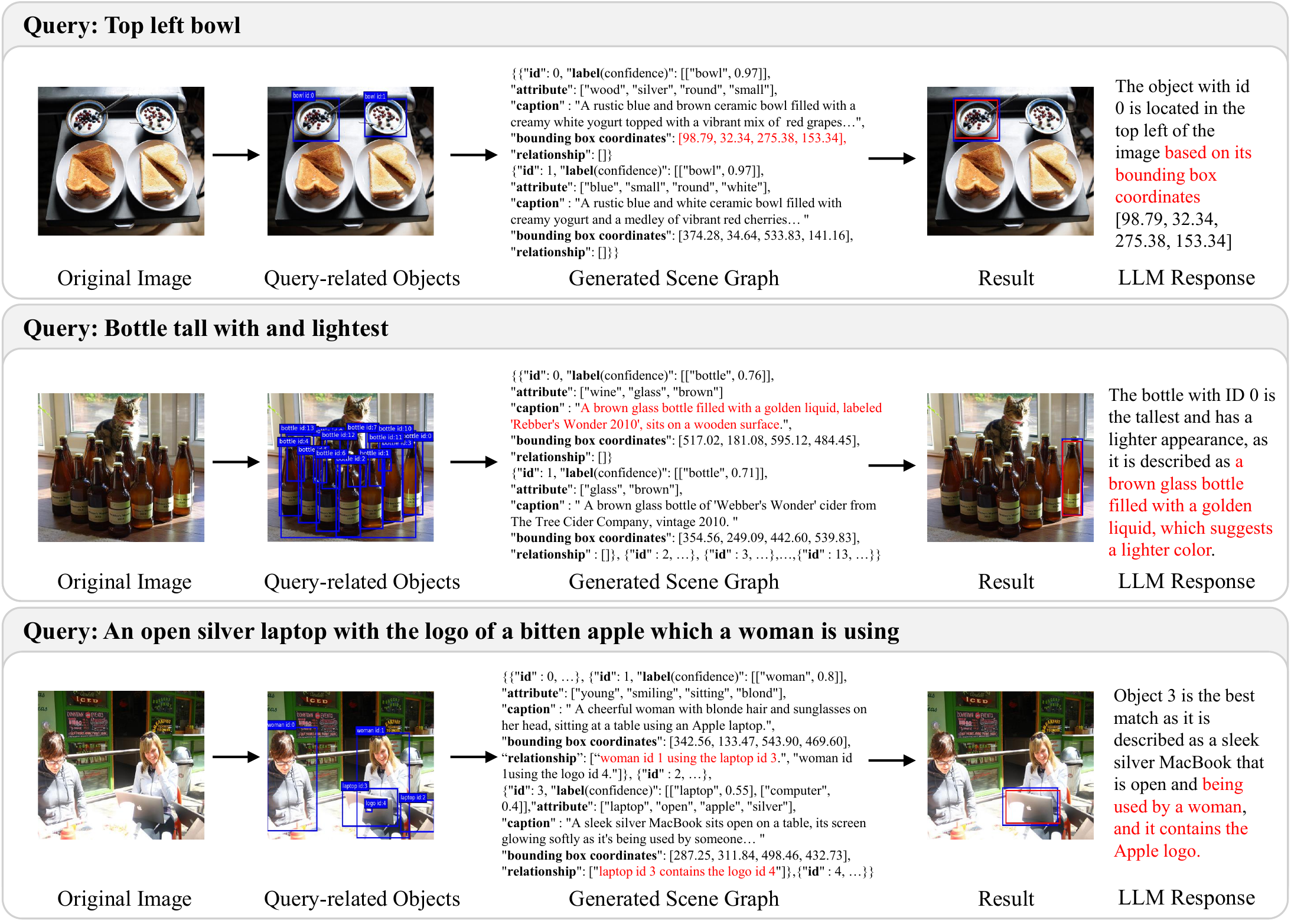}
 \caption{Correct comprehension of the detected targets and LLM explanations on RefCOCO, RefCOCO+, and RefCOCOg. The \textcolor{red}{red} bounding boxes indicate the ground truth annotations, and the \textcolor{blue}{blue} bounding boxes depict our predicted results.}
 \label{fig_vis}
\end{figure*}

\begin{figure*}[t!]
    \centering
    \includegraphics[scale=0.32]{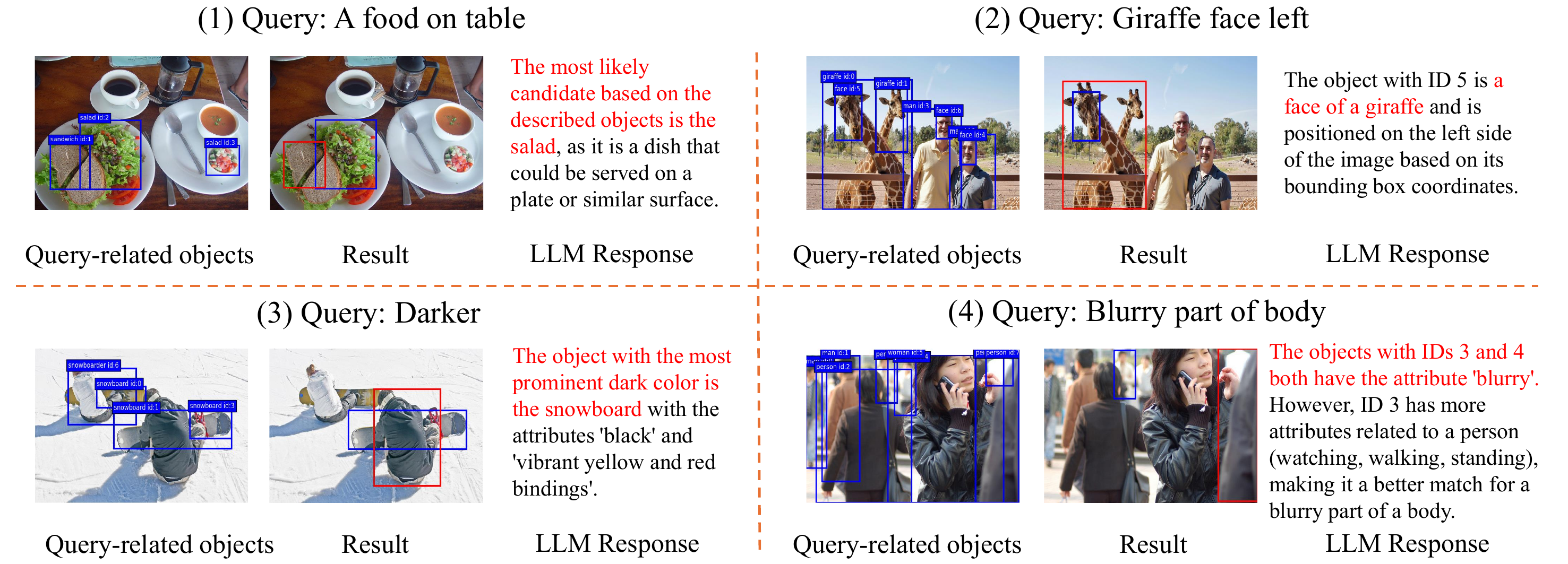}
 \caption{Failure case visualization. The \textcolor{red}{red} bounding boxes indicate the ground truth annotations, and the \textcolor{blue}{blue} bounding boxes depict our predicted results.
 }
 \label{fig_failure}
\end{figure*}

\subsection{Error Analysis}
We conduct additional analysis to investigate the error contributions of each stage and identify several common sources of failure beyond ambiguous queries.

\begin{table}[]
\caption{Analysis of error types within the framework. Reported rates (\%) correspond to averages over the sub-splits of each dataset.
 }
 \label{error}
\begin{tabular}{l|ccc}
\hline
Error rate                 & RefCOCO & RefCOCO+ & RefCOCOg \\ \hline
Missed detection           & 10.9\%  & 12.2\%   & 8.6\%    \\
Labeling error rate~\cite{chen2025revisiting} & 14\%    & 24\%     & 5\%      \\
LLM misinterpretation      & 7.6\%   & 6.4\%    & 13.2\%   \\ \hline
\end{tabular}
\end{table}

\noindent \textbf{Object detection failures} To investigate the errors originating from object detection, we measure the percentage of ground-truth bounding boxes that are covered by the candidate boxes produced by the detector, after applying our filtering strategy. As shown in the following table, the candidate bounding boxes include 89.12\% of the ground-truth boxes on RefCOCO val, 88.34\% on RefCOCO+ val, and 91.14\% on RefCOCOg val. This illustrates that nearly 10\% of the prediction error arises from the object detection stage.

\noindent \textbf{Labelling errors} As reported in Ref-L4~\cite{chen2025revisiting}, a non-negligible portion of RefCOCO(+/g) queries suffer from annotation inconsistencies, where the ground-truth boxes do not match the query semantics. We cite these statistics to account for part of the performance ceiling.

\noindent \textbf{LLM misinterpretation} Based on remaining cases, we estimate that roughly 10\% of errors stem from LLM misinterpretation. This typically occurs when scene graphs become too crowded or under-informative. For example, given 10 nearly identical objects and a query like “bottom left second from bottom,” the LLM may fail to disambiguate based solely on the scene graph description.

\subsection{Qualitative Results}
\label{qualitative}
\noindent \textbf{Correct Comprehension of Detection Results} 
Figure~\ref{fig_vis} illustrates the correct detection results, generated scene graphs, and corresponding LLM explanations for selected examples across three datasets. It can be seen that SGREC can infer spatial relationships by computing box coordinates, analyzing object appearance from image captions, and identifying complex interactions through the generated scene graphs. Moreover, for queries with ambiguous meanings (e.g., ``Bottle tall with and lightest"), SGREC can still localize the correct object by deducing that a ``lighter color" corresponds to a ``golden liquid." This demonstrates the robustness of SGREC, which harnesses strong language understanding capabilities in LLMs by reframing the image comprehension process as a textual reasoning task.

\noindent \textbf{Failure Case Analysis} 
SGREC encounters difficulties when the input query contains semantic ambiguity or fails to refer to the specific object in the image, which negatively impacts both language interpretation and object localization. As illustrated in Figure~\ref{fig_failure}, example (1) shows that the query "A food on table" is challenging due to multiple candidate food items, making it hard to determine the correct target. In example (2), the query "Giraffe face left" is ambiguous. It is unclear whether it refers to a giraffe facing left or the left side of a giraffe’s face. Similar issues appear in examples (3)–(4), where unclear queries prevent the model from reliably identifying the correct object.

\subsection{Limitation}
SGREC demonstrates strong zero-shot performance on the RefCOCO/+/g datasets but introduces additional computational costs due to involving two large models: subgraph generation via VLMs and final inference via LLMs. This trade-off enables the use of various pre-trained VLMs and LLMs without fine-tuning, but also relies on scene graphs to provide structured representations of the input image. Future work could explore unified VLM architectures that combine object detection and scene graph construction to enhance scalability and robustness.


\section{Conclusion}
We introduce SGREC, a novel zero-shot REC method that uses LLMs for object localization with scene graphs. First, we identify relevant objects by matching labels with queries’ extracted nouns, predicted categories, and inferred subjects. Next, we generate a scene graph for each image by integrating spatial information, object captions, and object interactions to describe the visual context. Finally, we localize the target object by reasoning over both the scene graphs and queries using LLMs. Experimental results show that SGREC achieves leading performance across most splits. SGREC bridges semantic gap between visual and text by leveraging scene graph representations of images.

\bibliography{egbib}
\bibliographystyle{IEEEtran}

\vfill

\end{document}